\PassOptionsToPackage{table}{xcolor}
\documentclass[10pt,twocolumn,letterpaper]{article}

\usepackage[pagenumbers]{cvpr} 
\usepackage{lipsum}
\usepackage{multicol}

\usepackage{fixmath}
\usepackage{bm}
\usepackage{epsfig}
\usepackage{graphicx}
\usepackage{amsmath}
\usepackage{mathtools}
\usepackage{amssymb}
\usepackage{latexsym}

\usepackage{verbatim}
\usepackage{algorithm}
\usepackage{algorithmic}
\usepackage{enumitem}
\usepackage{booktabs}
\usepackage{color, xcolor}

\newcommand\blfootnote[1]{%
  \begingroup
  \renewcommand\thefootnote{}\footnote{#1}%
  \addtocounter{footnote}{-1}%
  \endgroup
}

\usepackage[dvipsnames]{xcolor}

\definecolor{cvprblue}{rgb}{0.21,0.49,0.74}
\usepackage[pagebackref,breaklinks,colorlinks,citecolor=cvprblue]{hyperref}

\def\paperID{} 
\def\confName{CVPR}
\def\confYear{2026}

\title{Towards Highly-Constrained Human Motion Generation with \\ Retrieval-Guided Diffusion Noise Optimization }

\author{Hanchao Liu$^{1}$ \quad Fang-Lue Zhang$^{2}$ \quad Shining Zhang$^{1}$ \quad Tai-Jiang Mu$^{1\dag}$ \quad Shi-Min Hu$^{1}$ \\
$^1$ Tsinghua University \quad $^2$ University of New South Wales}

\begin{document}

\maketitle

\begin{abstract}
Generating human motion that satisfies customized zero-shot goal functions, enabling applications such as controllable character animation and behavior synthesis for virtual agents, is a critical capability. While current approaches handle many unseen constraints, they fail on tasks with very challenging spatiotemporal restrictions, such as severe spatial obstacles or specified numbers of walking steps. To equip motion generators for these highly constrained tasks, we present a retrieval-guided method built on the training-free diffusion noise optimization framework. The key idea is to search within large motion datasets for guidance that can potentially satisfy difficult constraints. We introduce relational task parsing to group target constraints and identify the difficult ones to be handled by retrieved reference. A better initialization for diffusion noise is then obtained via a reward-guided mask that combines random noise with retrieved noise. By optimizing diffusion noise from this improved initialization, we successfully solve highly constrained generation tasks. By leveraging LLM for relational task parsing, the whole framework is further enabled to automatically reason for what to retrieve, improving the intelligence of moving agents under a training-free optimization scheme. 
Project page: https://hanchaoliu.github.io/RetrievalGuidedDNO/
\blfootnote{$^{\dag}$ Corresponding author. {\tt {taijiang}@tsinghua.edu.cn} }

\end{abstract}    
\section{Introduction}
\label{sec:intro}

Generating human motion with diverse real-world controls—beyond textual descriptions and including interaction with surrounding scenes \cite{pan2025tokenhsi, wan2024tlcontrol, xie2023omnicontrol}—is receiving increasing attention. These control tasks require a higher level of generation capability and are critical for embodied agents and robotics in the physical world \cite{xu2024humanvla,chen2025gmt}. 
Recently, interpreting task goals as constraints and utilizing training-free optimization \cite{karunratanakul2023guided, liu2024programmable, karunratanakul2024optimizing, pinyoanuntapong2024controlmm} has proven effective for handling zero-shot spatial and temporal constraints. In this paper, we focus specifically on human motion generation within this training-free framework.

\begin{figure}[!t]
  \centering
   \includegraphics[width=0.98\linewidth]{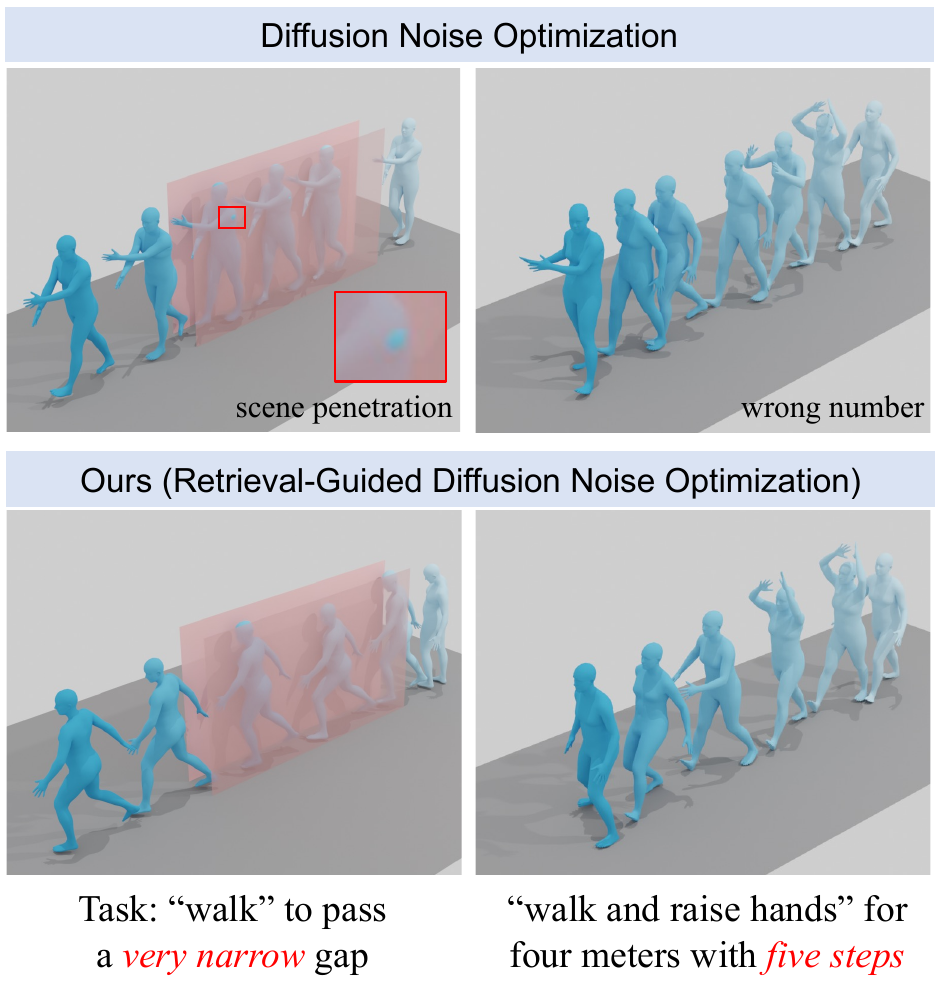}
   \caption{\textbf{Training-free Human Motion Generation for Highly-constrained Generation Tasks.} Compared to existing diffusion noise optimization methods \cite{liu2024programmable, karunratanakul2024optimizing} which exhibit high constraint error and motion artifacts, our proposed Retrieval-Guided Diffusion Noise Optimization significantly improves performance on these tasks. This improvement is achieved by leveraging relevant skills retrieved from existing motion datasets.}
   \label{fig:teaser}
   \vspace{-5pt}
\end{figure}

The motion programming framework \cite{liu2024programmable} and diffusion noise optimization (DNO) \cite{karunratanakul2024optimizing} allow for constructing arbitrary constraint functions and handling a wide range of unseen tasks. However, we observe that their capabilities remain limited when constraints become highly challenging. For example, as illustrated in Fig.~\ref{fig:teaser}, for controlling the precise number of steps while walking a specific distance or walking through a narrow gap, they yield motions with high constraint error and poor quality. We aim to address these \textit{highly-constrained} motion generation tasks, which typically involve (1) difficult spatiotemporal constraints and (2) behavioral constraints based on numerical control. These tasks are critical for practical applications, as real-world environments impose \textit{strict} requirements that must be met. To date, few works have specifically addressed this challenge or offered effective solutions.

We hold the key insight that the diffusion noise in the early stage matters for good generation. Therefore, we propose initializing the diffusion process with noise that is more suitable for satisfying the imposed constraints. While random noise can capture general constraints and textual semantics, challenging constraints require specific knowledge beyond its capacity. We therefore leverage existing datasets as a knowledge base from which to retrieve appropriate reference motion skills. As shown in Fig.~\ref{fig:teaser}, retrieval can identify noise patterns better suited to navigating narrow gaps (e.g., side walking) or performing a specific number of steps. The initial random noise is then guided towards this reference noise, gradually incorporating relevant motion skills retrieved from the dataset for the given task.

Motivated by the above, we propose a novel \textit{Retrieval-Guided Diffusion Noise Optimization} (RG-DNO) framework for generating human motion. RG-DNO extends the programmable motion generation framework \cite{liu2024programmable} by addressing the question of how to parse and handle constraints when they are particularly challenging. 
Given the set of constraints describing a motion generation task, we first conduct relational task parsing, i.e, to identify a constraint subset for performing retrieval based on the difficulty ordering and relations among the required constraints. An initial noise is then optimized via a masked composition of a random noise and the retrieved noise, based on its ability to satisfy the complete set of constraints and motion quality rewards. Taking it as improved initialization, we optimize for the final motion result. Notably, the relational task parsing can also be made automatic by leveraging the strong reasoning ability of a large language model (LLM).

We designed a set of challenging motion generation tasks involving complex scene interactions and numerical control for validation. The experiments demonstrate that our method successfully solves these constraint-based generation tasks with low error and high motion quality, outperforming the native DNO approach. Our method can be readily integrated with standard diffusion noise optimization to improve constraint satisfaction beyond what existing approaches can achieve, enabling new control capabilities for embodied agents within a training-free framework.

In summary, the contributions of this work include:

\begin{itemize}
    \item We are the first to address highly-constrained motion generation tasks, identifying the limitations of current diffusion noise optimization in handling difficult constraints.
    \item We propose a guided initial noise optimization approach that leverages knowledge of relevant skills retrieved from existing large-scale datasets. We formulate this initial noise construction as a novel framework with relational task parsing, retrieval, and masked noise optimization.
    \item We introduce a set of zero-shot generation tasks that extend constraints to body shape and fine-grained motion control (e.g., numbers of steps), providing a benchmark for evaluating highly-constrained generation and demonstrating promising results. 

\end{itemize}

\section{Related Work}
\label{sec:related}

\textbf{Human Motion Generation and Control.} 
Text-to-motion generation has attracted wide attention, including diffusion models \cite{tevet2022human,zhang2024motiondiffuse,chen2023executing}, auto-regressive models \cite{jiang2023motiongpt,fan2025go,huang2024controllable}, and hybrids of both \cite{zhaodartcontrol}. Two types of control signals are commonly used for controllability: trajectory signals and constraint functions. Trajectory signals \cite{shafir2023human,xie2023omnicontrol,wan2024tlcontrol,hou2024causal} can serve as conditions for ControlNet-like architectures but struggle with precise scene/object interaction tasks. 
Constraint functions \cite{liu2024programmable,karunratanakul2024optimizing,karunratanakul2023guided} offer a flexible representation for arbitrary zero-shot generation tasks. 
They are readily compatible with training-free diffusion noise optimization \cite{liu2024programmable,karunratanakul2024optimizing} and can also integrate trajectory signals as optimization goals \cite{liu2024programmable,pinyoanuntapong2024controlmm}. In this work, we adopt the latter approach for customized motion generation.
Recently, preference learning and fine-tuning methods \cite{han2025atom,wangaligning,tan2024sopo} have been proposed to provide finer textual control over action frequency \cite{han2025atom,tan2024sopo} and amplitude \cite{huang2024controllable,li2025simmotionedit}; however, they do not support customized constraint functions.

\begin{figure*}[!t]
  \centering
   \includegraphics[width=\linewidth]{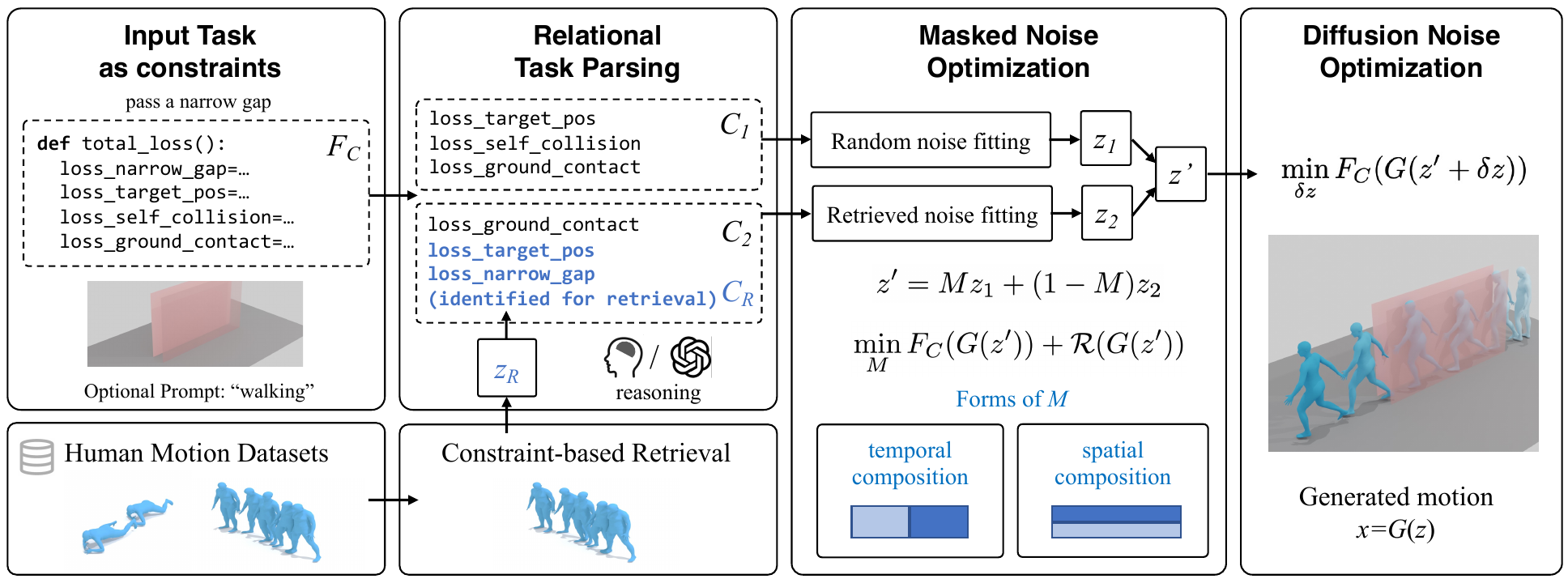}
   \caption{\textbf{Overview of Retrieval-Guided Diffusion Noise Optimization.} Given a motion generation task represented by a combined constraint function $F_C$, we apply either manual or LLM-based relational task parsing to group difficult constraints $C_R$ for retrieving potential skills $x_R$, and to group the remaining constraints into subsets $C_1$ and $C_2$ that can be handled respectively by random and retrieved noises. Using a motion diffusion model $G$, we optimize noises $z_1$ and $z_2$ to satisfy $C_1$ and $C_2$, starting from random noise and from the retrieved noise $z_R$. We also optimize a linear mask $M$ to combine them into $z'$ based on the constraint error $F_C$ and a quality-check reward $\mathcal{R}$. The final motion is produced by using $z'$ as an improved initialization and performing a final round of diffusion noise optimization.}
   \label{fig:main_overview}
\end{figure*}

\noindent\textbf{Retrieval-based Approaches for Human Motion Generation.} 
How to make full use of ready-to-use motion samples in existing datasets is a long-standing question. A number of retrieval-, matching-, and composition-based approaches have been proposed for human motion generation tasks \cite{zhang2023remodiffuse,li2023example,athanasiou2023sinc,mandelli2025generation,petrovich2024multi}. 
Generative motion matching \cite{li2023example} can produce long motions, but it lacks controllability. Retrieval augmentations such as ReMoDiffuse \cite{zhang2023remodiffuse}, ReMoMask \cite{li2025remomask}, RMD \cite{liao2024rmd}, and video augmentation methods \cite{xu2025vimorag} have been integrated with generation models to improve performance. SIMS \cite{wang2024sims} generates long-horizon scripts for motion generation using RAG. 
The pose-guided diffusion model \cite{caipose} also uses poses as references for unseen motions but cannot handle constraints. We first adopt retrieval to provide potential motion skills for constrained motion generation.

\noindent\textbf{Noise Optimization for Diffusion Models.} 
For diffusion-based text-to-image models, many studies optimize latent diffusion noise as an effective way to perform content editing \cite{xie2025cannyedit,cao2023masactrl,pham2025autoedit,wang2025diffusionsurvey} and search for better initial noises to improve alignment with complex textual conditions \cite{eyring2024reno,miao2025noise}. 
Several works highlight the relationship between the sampled initial noise and the inherent structure of the final generated results \cite{guo2024initno,lee2025countcluster,qi2024not,xu2025good}. 
Our work shares the same motivation that the choice of initial noise is important, but we address the limitations of random noise for handling difficult constraints and leverage guidance from retrieved samples to craft better diffusion noise.

\section{Preliminaries}

\noindent\textbf{Motion Diffusion Models.} 
A Motion Diffusion Model (MDM) \cite{tevet2022human} iteratively denoises a random noise from $z \in \mathbb{R}^{T\times d} \sim \mathcal{N}(0,I)$ to a motion sequence with $T$ frames and pose dimension $d$ with optional text condition. DDPMs inject fresh random noise at each denoising step, whereas DDIMs \cite{song2020denoising} rely only on the initial noise according to:
\begin{equation}
    z_{t-1} = \sqrt{\frac{\alpha_{t-1}}{\alpha_t}}(z_t - \sqrt{1-\alpha_t} \epsilon_\theta(z_t)) + \sqrt{1- \alpha_{t-1}}  \epsilon_\theta(z_t)
\end{equation}
Although the denoising process can be performed in a latent feature space using an additional VAE \cite{chen2023executing}, in this work, we use the native MDM for its simplicity.

\noindent\textbf{Mesh Representation for Human Motion.} 
The motion representation used by native MDM \cite{guo2022generating} cannot directly yield SMPL \cite{loper2015smpl} parameters and the corresponding mesh, which hinders direct interaction between the body surface and the surrounding scene. To address this, we adopt the RoHM motion representation \cite{zhang2024rohm} and remove unnecessary body-shape features. Each frame's pose is thus represented by a 284-dimensional feature $(\mathbf{r}, \mathbf{p})$, where $\mathbf{r}$ is the root trajectory and $\mathbf{p}$ is the pose feature. Specifically,
$\mathbf{p} = (\mathbf{J}, \dot{\mathbf{J}}, \boldsymbol{\theta}, \mathbf{f})$,
where $\mathbf{J}$ are joint positions, $\dot{\mathbf{J}}$ are joint velocities, $\boldsymbol{\theta}$ are SMPL rotations, and $\mathbf{f}$ are foot-contact labels.

\noindent\textbf{Programmable Motion Generation and Noise Optimization.} Both ProgMoGen \cite{liu2024programmable} and Diffusion Noise Optimization (DNO) \cite{karunratanakul2024optimizing} were proposed concurrently to apply diffusion noise optimization for human motion control. ProgMoGen introduces a motion programming framework to construct combined constraint functions for arbitrary generation tasks; it samples random noise and optimizes it using Adam optimizer \cite{kingma2014adam}. DNO introduces gradient normalization and learning rate decay strategies to improve convergence and is primarily used for motion editing and refinement given a reference sample. In this paper, we combine elements from both methods for customized generation tasks.
Specifically, given a task defined by a goal function $F$, we sample random noise $z\sim \mathcal{N}(0,I)$, and optimize
\begin{equation}
    \min_{z} F(G(z,\mathcal{C}))
    \label{eq:main_min}
\end{equation}
where $G$ is the frozen generator (e.g., an MDM) and $\mathcal{C}$ is the text condition. In each optimization step, the gradient is normalized as $\nabla_z F \xleftarrow{ }\nabla_z F / \parallel \nabla_z F  \parallel $. We also adopt cosine learning rate decay with warm-up.

\section{Method}
\label{sec:method}

\subsection{Overview} 

The aim of this work is to address highly constrained human motion generation tasks within a training‑free diffusion noise optimization framework. Given a generation task specified by a combined constraint function $F_C$ defined over a constraint set $C$ (obtainable via motion programming \cite{liu2024programmable}), our goal is to generate a motion sequence $x\in\mathbb{R}^{T\times d}$. This problem can be formulated as:
\begin{equation}
    \min_z \sum_i F_{C_i}(G(z,C_0))
\end{equation}
Here, $G$ is a pre-trained generator and $C_0$ is an optional text condition. As shown in Fig.~\ref{fig:main_overview}, we first perform relational task parsing (Sec.~\ref{sec:task_partitioning}) to group constraints for retrieval and for each optimization phase of the pipeline. The constraint-based retrieval (Sec.~\ref{sec:constraint_retrieval}) searches a motion dataset $\mathcal{D}$ for samples that may satisfy the hard constraints. We then optimize two diffusion noise initializations—one from random noise and one from the retrieved noisy latent—to fit the parsed constraint subsets. A linear mask $M$ is optimized to combine these two noises into $z'$ using retrieval-based and reward-based guidance (Sec.~\ref{sec:noise_opt}). Finally, $z'$ is used to initialize a final diffusion noise optimization that produces the output motion. Details of the mask $M$ and its optimization are given in Sec.~\ref{sec:noise_opt_details}.

\subsection{Relational Task Parsing} 
\label{sec:task_partitioning}

Following a divide‑and‑conquer strategy, we first decide which constraints in the full set should be addressed by retrieval and which should be handled by noise‑based optimization. Specifically, we propose \textit{relational task parsing} to address the above question by determining the difficulty ordering and relationships among constraints in $C$. Formally, the goal is to identify a subset $C_R$ for retrieval and group the remaining constraints into $C_1$ and $C_2$, to be optimized separately with random noise and with retrieved noise, respectively, before merging them:
\begin{equation}
    C = C_1 \oplus C_2, C_R \subseteq C_2
\end{equation}
where $\oplus$ is the union operation and $C_1$ may also include textual semantics $C_0$. 

Intuitively, the most difficult constraints should be handled by retrieval, while constraints in $C_1$ and $C_2$ should yield motions that remain visually plausible to facilitate later composition. Accordingly, we adopt the following reasoning rules for task parsing:

\noindent\textbf{Rule 1}: Both sets $C_1$ and $C_2$ are encouraged to satisfy their constraints as fully as possible to reduce the difficulty of producing a coherent result during combination.

\noindent\textbf{Rule 2}: If a difficult constraint $c_D$ is identified, it should be included in the retrieval set $C_R$ together with its tightly connected constraints.

\noindent\textbf{Rule 3}: Constraints that may conflict with $C_R$ should be removed from $C_2$ to ensure high-quality motion generation for $C_2$. For example, a constraint requiring very low head height in the middle frame and high head height at the final frame would be considered conflicting.

\noindent\textbf{Rule 4}: If the retrieval is not fully confident, difficult constraints may optionally be included in $C_1$ and tackled using random noise as well.

Based on the rules above, we propose a greedy strategy (Algorithm~\ref{alg:task_parsing}) that takes the constraint set $C$ and a user-specified retrieval confidence score $s$ as input. The procedure is as follows: (1) initialize $C_1=C_2=C$; (2) identify the most difficult constraint $c_D$ from difficulty ordering and parse its relationships with other constraints; and (3) iteratively add or remove constraints from the retrieval set $C_R$ and from the fitting sets $C_1$ and $C_2$. Note that it is acceptable for $C_1=C_2=C$; in that case, we address the full problem from both the random- and retrieved-noise perspectives.

The proposed relational task parsing extends motion programming \cite{liu2024programmable} by further investigating the relationship between constraints. 
The difficulty ordering and relationship between constraints can be specified manually by end users for precise control, or inferred automatically via an LLM.

\renewcommand{\algorithmicrequire}{\textbf{input:}\unskip}
\renewcommand{\algorithmicensure}{\textbf{output:}\unskip}
\begin{algorithm}[!t]
  \caption{Relational Task Parsing}
  \label{alg:task_parsing}
  \small
  \begin{algorithmic}
    \REQUIRE Constraint set $C$, retrieval confidence $s \in \{0,1\}$.
    \ENSURE retrieval set $C_R$, random noise fitting set $C_1$, retrieval noise fitting set $C_2$.
    \STATE Initialize $C_1=C_2=C$, $C_R = \{\}$.
    \STATE Identify the most difficult constraint $c_{D}$. 
    \STATE Identify relation $e_k$ between $c_D$ and other constraint $c_k \in C$ . $e_k \in \{\text{`connected', `conflict', `none'}\}$.
    \STATE Add $c_D$ and $c_D.\text{connected()}$ to $C_R$. 
    \STATE Remove $c_D$.\text{conflict()} from $C_2$.
    \STATE if $s=1$, then remove $c_D$ from $C_1$.
  \end{algorithmic}
\end{algorithm}

\subsection{Constraint-based Retrieval} 
\label{sec:constraint_retrieval}
After identifying the retrieval set $C_R$, we aim to find corresponding motion samples from an existing dataset $\mathcal{D}$ that can potentially satisfy it. This is formulated as a retrieval problem: find a sample that yields the smallest expected error for the constraint set $C_R$.
In this way, we have:
\begin{equation}
    x = \arg \min_{x \in \mathcal{D}} F_{C_R}(x)
\end{equation}
\noindent\textbf{Equivalent Transform.} We allow motion samples to perform a spatial transform in the horizontal plane to fit the required constraints:
\begin{equation}
    x, \mathcal{H} = \arg\min_{x,\mathcal{H}} F_{C_R}(\mathcal{H}x)
\end{equation}
where $\mathcal{H}$ is a horizontal transformation.

\noindent\textbf{Semantic Consistency Check.} We filter out motions with significant semantic discrepancies by comparing the target text prompt to the sample annotations in $\mathcal{D}$.

\noindent\textbf{Temporal Resizing.} Retrieved samples often have different durations than the target; we resample them to the target length using linear interpolation.

We reserve a top-$k$ set of retrieved samples $x_R$ whose constraint error falls below a threshold, and select one as $x_R$. We then fit a diffusion noise to imitate the chosen retrieved sample:
\begin{equation}
    z_R = G^{-1} (\mathcal{H} x_R, C_0)
\end{equation}
In this way, the diffusion noise $z_R$ can serve as a guidance for handling the difficult constraint.

\subsection{Masked Noise Optimization} 
\label{sec:noise_opt}

In this stage, we update the diffusion noise by combining random noise with the retrieved reference. The text condition $C_0$ is omitted here for brevity; it will be provided to the generation model if it exists. Starting from random noise $z_0\sim\mathcal{N}(0,I)$, we optimize diffusion noise $z_1$ to satisfy the constraint set $C_1$:
\begin{equation}
    \min_{z_1} F_{C_1} (G(z_1,z_0))
\end{equation}
Similarly, starting from the retrieved noise $z_R$, we optimize diffusion noise $z_2$ to fit the constraint set $C_2$:
\begin{equation}
    \min_{z_2} F_{C_2} (G(z_2,z_R))
\end{equation}
To combine noises $z_1$ and $z_2$ while incorporating the retrieved skills, a simple guidance is to move linearly towards $z_2$, yielding the updated noise $z'$ as:
\begin{equation}
    z' = M z_1 + (1-M) z_2
\end{equation}
Here $M$ is a mask with the same shape as $z_1$ and $z_2$, i.e., $M\in\mathbb{R}^{T\times d}$. 
We aim to find $M$ such that the combined noise $z'$ satisfies the full constraint set $C$ while preserving motion quality. 
Accordingly, $M$ is optimized as:
\begin{equation}
\label{eq:opt_mask}
    \min_{M} F_{C} (G(z')) + \mathcal{R} (G(z'))
\end{equation}
where $\mathcal{R}$ is a reward function in order to achieve a good trade-off between constraint error and motion quality. The type of reward function $\mathcal{R}$ and the solution to this problem will be discussed later.

After obtaining $M$ and $z'$, we start from this new initialization $z'$ and perform another round of standard diffusion noise optimization:
\begin{equation}
\label{eq:opt_final}
    \min_{\delta z} F_{C} (G(z'+\delta z))
\end{equation}
The final result is obtained as $x=G(z'+\delta z)$.

\subsection{Reward-Guided Mask Optimization} 
\label{sec:noise_opt_details}

Directly optimizing $M$ in Eq.~(\ref{eq:opt_mask}) yields a very large search space and can degrade the motion priors encoded in $z_1$ and $z_2$. To reduce complexity, we adopt a heuristic mask selection strategy. We construct a set of downsampled binary mask candidates $\mathcal{M}$ that vary along temporal and spatial (pose‑feature) dimensions, i.e., $\mathcal{M}=\{{M|M_{ij}\in \{0,1\}}\}$. We then select the best mask from $\mathcal{M}$ by minimizing the target objective:
\begin{equation}
    M = \arg \min_{M\in \mathcal{M}} F_{C} (G(z')) + \mathcal{R} (G(z'))
    \label{eq:mask_candidate}
\end{equation}
This approximate mask solution is tractable and reasonable because it undergoes a subsequent optimization round to fit $F_C$, and it enables flexible spatial and temporal composition across different generation tasks. The choice between temporal and spatial masks can be made according to Eq.~(\ref{eq:mask_candidate}) or specified based on the generation task.

\noindent\textbf{Temporal Mask.}
To construct mask candidates, we split $M$ into $N_T$ temporal segments, and set every entry within the $i$-th segment $M[T/N_T*i:T/N_T*(i+1),:]$ uniformly to either $0$ or $1$. This yields $2^{N_T}$ mask candidates (e.g., $N_T=8$ in practice). We evaluate all candidates and select the one that minimizes the error. To refine sharp boundaries, we further optimize a soft blending using a sigmoid representation (See the supplementary material).

\noindent\textbf{Spatial Mask.} 
Similarly, we split the mask into $N_S$ parts along the pose‑feature dimension and set all entries in the $i$‑th part uniformly to either $0$ or $1$, yielding $2^{N_S}$ mask candidates. In practice, we partition pose features into root trajectory, left/right arm and hand, left/right leg, head, and two spine segments ($N_S=8$).

\noindent\textbf{Reward Function.} The motion-quality reward function $\mathcal{R}$ helps filter out spatial or temporal compositions that cause large inconsistencies between frames or between body parts, which would otherwise be passed to the final optimization stage. We use an aggregated reward composed of several simple motion‑check cost functions $\mathcal{L}_k$:
\begin{align}
    \mathcal{R}(G(z'), z') =\lambda_1 & \mathcal{L}_{\text{jitter}}(G(z')) + \lambda_2 \mathcal{L}_{\text{foot skate}}(G(z')) + \nonumber \\ 
    \lambda_3 & \mathcal{L}_{\text{decorr}}(z') + 
    \lambda_4 \mathcal{L}_{\text{semantic}}(G(z')) 
    \label{eq:reward_func}
\end{align}
$\mathcal{L}_k$ includes joint-based maximum jitter and foot skating to punish inconsistent masked combinations between frames and body parts, and can optionally include noise de-correlation loss \cite{karunratanakul2024optimizing} and semantic alignment score, e.g., motion-text similarity \cite{petrovich2023tmr,karunratanakul2024optimizing}. The mask selection process is fast, as the constraint error and reward calculation for each candidate are easy to compute and can be performed in parallel. 
The weights $\lambda_k$ are set to $1.0$ as a default and can be adjusted to fit specific tasks.

\section{Experiments}
\label{sec:exp}

\setlength{\tabcolsep}{2pt}
\begin{table*}[!t]
    \centering
    \small
    \begin{tabular}{l|cccc|cccc} 
  \toprule
    & \multicolumn{4}{|c|}{Task-1: very narrow gap} & \multicolumn{4}{c}{Task-2: very low barrier} \\
    \midrule
    Method           & Foot Skate $\downarrow$  & Max Acc. $\downarrow$ & C. Error $\downarrow$ & Max SP. $\downarrow$ &  Foot Skate $\downarrow$ &  Max Acc. $\downarrow$ & C. Error $\downarrow$ & Max SP. $\downarrow$ \\  
   \midrule
   Unconstrained MDM-RoHM \cite{tevet2022human}   & 0.038	& 0.094 & 14.101 & 0.506 & 0.025  & 0.098 & 11.755 & 1.173  \\ 
   ProgMoGen+DNO \cite{liu2024programmable,karunratanakul2024optimizing}   & \textbf{0.029} & 	0.178 &	0.0162 & 0.073 & \textbf{0.221}  & 0.261 & 0.000115 & 0.149   \\ 
   
   Ours   &  0.040 &	\textbf{0.120} &	\textbf{0.0050} & \textbf{0.027} & 0.228 & \textbf{0.194} & \textbf{0.000049} & \textbf{0.079} \\ \midrule
   ProgMoGen+DNO($N_S$=5) \cite{liu2024programmable,karunratanakul2024optimizing}   &  \textbf{0.014} & 0.150 & 0.0019 & 0.029 &  \textbf{0.189} &  0.240 &  0.000015 & 0.064 \\ 
   Ours($N_S$=5)   & 0.039 & \textbf{0.109} & \textbf{0.0000} & \textbf{0.001} & 0.193	& \textbf{0.157} & \textbf{0.000007} & \textbf{0.048}  \\ 
  \bottomrule
    \end{tabular}
    \vspace{-4pt}
    \caption{Quantitative comparison on the highly-constrained motion generation tasks involving challenging spatiotemporal constraints. $N_S$=5 denotes a 5-run initial noise search.}
    \label{tab:1}
\end{table*}
\setlength{\tabcolsep}{1.4pt}

\setlength{\tabcolsep}{5.5pt}
\begin{table*}[t]
    \centering
    \small
    \begin{tabular}{l|cccccc} 
  \toprule
    & \multicolumn{6}{c}{Task-3: Assign number of walking steps \textit{with action: raise hand}} \\
    \toprule
    Method           & Foot Skate $\downarrow$ & Max Acc. $\downarrow$ & C. Error $\downarrow$ & Succ. Rate $\uparrow$ &   Sem. Succ. Rate $\uparrow$ & Pace Pattern $\uparrow$ \\  
   \midrule
   Unconstrained MDM-RoHM \cite{tevet2022human}   & 0.004 & 0.074 & 18.6 & 0.031 & 0.031 & 0.969  \\ 
   ProgMoGen+DNO \cite{liu2024programmable,karunratanakul2024optimizing}   & 0.023 & 0.116 & 0.282 & 0.469 & 0.375 & 0.843  \\ \midrule
   Ours   & \textbf{0.021} & \textbf{0.115} & \textbf{0.0003} & \textbf{0.594} & \textbf{0.438} & \textbf{0.875}  \\ 
  \bottomrule
    \end{tabular}
    \vspace{-4pt}
    \caption{Quantitative comparison on the highly-constrained motion generation task involving challenging numerical constraints.}
    \label{tab:2}
\end{table*}
\setlength{\tabcolsep}{1.4pt}

\subsection{Tasks for Evaluation}
Highly-constrained motion generation remains relatively unexplored in the existing literature. In order to evaluate our method, we design three quantitative evaluation tasks. For challenging spatiotemporal constraints, following ProgMoGen \cite{liu2024programmable}, we let the character pass through barriers and enforce constraints on the fine-detailed SMPL mesh to more accurately reflect these constraints.  These include \textbf{Task-1:}\textit{walk through a very narrow gap} and \textbf{Task-2:}\textit{walk to avoid a very low overhead barrier}. To assess numerical constraint handling, we introduce \textbf{Task-3:}\textit{walk a precise number of steps to reach a target}, accompanied by other actions.

In Task-1, the width of the narrow gap is set to 0.4 meters, and the walking distance is 5 meters. 
In Task-2, the height of the overhead barrier is set to 0.5 meters between $2<z<3$ and the walking distance is set to 5 meters. In Task-3, the number of walking steps is set to 6 for walking 4 meters, together with the action \textit{raising hand}. 
Additional constraints are also included as regularization terms. 
The numerical constraints are implemented as differentiable counting functions.
The complete set of constraints and the loss functions are provided in the supplementary material.

\noindent\textbf{Evaluation Metrics.}
Following ProgMoGen \cite{liu2024programmable}, we assess the generated motions using: the foot skate ratio (Foot Skate) to measure motion coherence, maximum joint acceleration (Max Acc.) to quantify joint jitter and inter-frame motion consistency, and constraint error (C.Err) to evaluate constraint satisfaction. 
The task-dependent maximum scene penetration (Max SP.) is also reported for Task-1 and Task-2.
For Task-3, we use the success rate (Succ. Rate), which denotes the percentage of trials where the assigned number of steps is followed accurately, and the semantic success rate (Sem. Succ. Rate) to assess the trials where both the assigned number of steps and the accompanying action are correctly performed. The pace score, which evaluates the consistency of the alternating stepping pattern, is also used. Note that the success rate and pace score are measured between 0 and 1, higher values indicating better performance.
Detailed descriptions for these metrics are provided in the supplementary material.

\subsection{Implementation Details}

\textbf{Datasets.}
We use the HumanML3D dataset \cite{guo2022generating} for training the base model as well as conducting constraint-based retrieval in our approach. HumanML3D contains 14,616 diverse motion samples mainly from the AMASS \cite{mahmood2019amass} dataset. We use the training split for training the base model, and use the whole dataset for retrieval.

\noindent\textbf{Base Model.}
We pre-process the HumanML3D dataset to the 284-dim RoHM motion representation \cite{zhang2024rohm} and re-train the Motion Diffusion Model (MDM) \cite{tevet2022human} with its default training configuration (denoted as MDM-RoHM). We use its DDIM model with 50 diffusion steps for generation, which can achieve a better balance between performance and efficiency as demonstrated in \cite{tevet2022human, karunratanakul2024optimizing}.

\setlength{\tabcolsep}{2.5pt}
\begin{table}[t]
    \centering
    \small
    \begin{tabular}{lccc} 
  \toprule
    \multicolumn{4}{c}{Task HSI-2: avoiding barrier \cite{liu2024programmable}} \\
    \toprule
   Method            & Foot Skate $\downarrow$  & Max Acc. $\downarrow$  & C.Err $\downarrow$ \\  
   \midrule
   Unconstrained MoMask \cite{guo2024momask}    & 0.072 & 0.117 & 0.464   \\ 
   MaskControl \cite{pinyoanuntapong2024controlmm}   & \textbf{0.146} & 0.126 & \textbf{0.000}  \\  \midrule
   Unconstrained MDM \cite{tevet2022human} & 0.096  & 0.126  & 0.454  \\ 
   ProgMoGen \cite{liu2024programmable} & 0.189 &  0.150 & 0.097    \\ 
   Ours   & 0.165 & \textbf{0.123} & \textbf{0.000}  \\
  \bottomrule
    \end{tabular}
    \vspace{-4pt}
    \caption{Quantitative comparison on Task HSI-2 with joint-based constraints. MaskControl uses MoMask as its base model.}
    \label{tab:3}
\end{table}
\setlength{\tabcolsep}{1.4pt}

\begin{figure*}[!t]
  \centering
   \includegraphics[width=\linewidth]{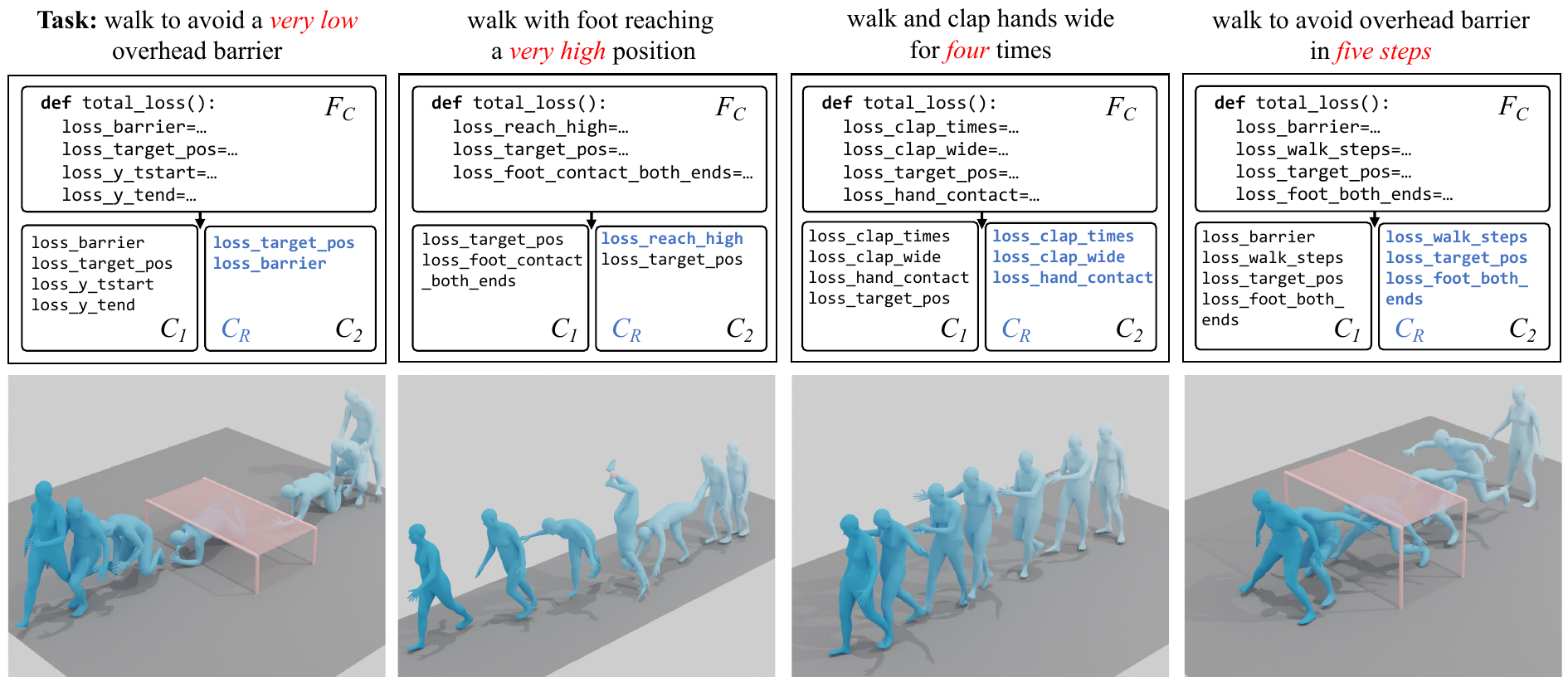}
   \vspace{-19pt}
   \caption{Qualitative examples for various highly-constrained generation tasks. The relational task parsing results are obtained via LLM. Details of each constraint function are provided in the supplementary material.}
   \label{fig:qualitative}
   \vspace{-8pt}
\end{figure*}

\noindent\textbf{Hyper-parameters.}
For our method, we use a learning rate of 0.05 and set the optimization step to $N_1 = 100$ for the first stage. This includes optimization of $z_1$, $z_2$, and $z'$, as well as fitting the transformed retrieved sample $x_R$. 
For optimization of $\delta z$, we reduce the learning rate to 0.02 and set the optimization steps to $N_2 = 400$.

\subsection{Baseline Methods}

We compare against state-of-the-art approaches ProgMoGen \cite{liu2024programmable} and MaskControl \cite{pinyoanuntapong2024controlmm}, as they support motion generation with customized constraint functions. We enhance ProgMoGen using the optimization strategies described in DNO \cite{karunratanakul2024optimizing}, and refer to this baseline as ``ProgMoGen+DNO''. For fair comparison, the base model for ProgMoGen+DNO is set to our re-trained MDM-RoHM. The learning rate is set to 0.05 and optimization is performed for 500 steps. MaskControl can only handle joint-based constraints, and we also compare against it on \textbf{Task HSI-2} from ProgMoGen \cite{liu2024programmable}, which uses joint-based constraints. The unconstrained model MDM-RoHM is a reference for performance when no constraints are imposed.
For evaluation, we generate 32 samples for each task.

\subsection{Quantitative and Qualitative Evaluation}

The results of our method and the baselines are presented in Tables \ref{tab:1} and \ref{tab:2}. Our method achieves significantly lower constraint error for tasks involving challenging spatiotemporal constraints. In addition, motion quality is also improved. Compared to ProgMoGen+DNO, the joint jitter is greatly reduced while the foot skating generally remains at a similar level. For tasks involving numerical constraints, our method better satisfies the constraints and achieves a higher success rate. Table \ref{tab:3} presents results for Task HSI-2 with joint-based constraints. Considering the performance difference of the base models that are used, our method and MaskControl yield comparable results.

We present qualitative examples for a diverse set of tasks in Fig.~\ref{fig:qualitative} with automatic relational task parsing result via an LLM, i.e., DeepSeek R1 \cite{guo2025deepseek} (see supplementary material for detailed LLM instruction). Our method retrieves appropriate skills such as \textit{side walking, crawling, hand standing} and succeeds in solving challenging tasks. As shown in Fig.~\ref{fig:compare_simple_retrieve} (a), the native ProgMoGen+DNO fails to find specific skills for \emph{foot reaching high}, resulting in physically unsatisfactory actions like \textit{stepping or floating in the air}.

\noindent\textbf{Comparison with Random Noise Search.}
Table \ref{tab:1} also shows results of incorporating a 5-run initial noise search ($N_S$=5) \cite{liu2024programmable}, where the result with the lowest constraint error is selected. We generally observe improved performance with this strategy. However, for the baseline method, while the noise search leads to a notably smaller constraint error, it does not substantially improve on joint jitter.

\subsection{Ablation Studies}

\setlength{\tabcolsep}{3.5pt}
\begin{table}[t]
    \centering
    \small
    \begin{tabular}{lcccc} 
    \toprule
    \multicolumn{5}{c}{Task-2: very low barrier} \\
    \toprule
   Method            & FS & Local FS  & Max Acc.  & C.Error \\  
   \midrule
   Random $z_1$ only               & 0.221 & 0.096 & 0.261 & \cellcolor{red!10}0.000115 \\ 
   Retrieval $z_2$ only            & 0.217 & \cellcolor{red!10}0.180 & 0.109 & 0.000014  \\ 
   \textit{w/o} task parsing $C_R$ & 0.221 & 0.065 & \cellcolor{red!10}0.289 & \cellcolor{red!10}0.000132  \\
   \textit{w/o} weight optim. $M$  & 0.195 & \cellcolor{red!10}0.306 & 0.123 & 0.000020  \\ 
   \textit{w/o} reward func. $\mathcal{R}$ & 0.240 & 0.108 & 0.202 & 0.000059  \\ 
   $C_1=C_2=C$                     & 0.235 & 0.150 & 0.169 & 0.000052  \\ \midrule
   Ours (full)                     & 0.228 & 0.134 & 0.194 & 0.000049  \\ 
  \bottomrule
    \end{tabular}
    \caption{Ablation study of the proposed method on Task-2. The red background indicates that the method fails in this metric.}
    \label{tab:ablation}
\end{table}
\setlength{\tabcolsep}{1.4pt}

\noindent\textbf{Effect of Each Module.}
To investigate the effect of each module in the proposed method, we perform an ablation study on Task-2 \emph{very low barrier}. We include the local foot skate ratio when $4<z<5$ to better reflect local motion quality. 
As shown in Table \ref{tab:ablation}, we observe the following: (1) Both random noise and retrieved noise alone are insufficient to generate high-quality results. Using only retrieved noise leads to poor local motion quality, such as excessive foot skating and implausible poses, as illustrated in Fig.~\ref{fig:compare_simple_retrieve} (b). (2) Retrieval based on the entire task does not yield good results, highlighting the necessity of our relational task parsing to identify the difficult set $C_R$. (3) Optimizing the mask $M$ is crucial; a simple linearly weighted combination (e.g., $M=0.5$ in this ablation) significantly reduces local motion quality, resulting in over-smoothing. (4) Incorporating the reward function enhances overall motion quality metrics. (5) A careful partitioning of tasks into $C_1$ and $C_2$ could potentially improve certain metrics, but may also introduce larger jitter, as this causes $z_1$ and $z_2$ to become less aligned before merging. This partitioning can be used to strike a balance between motion smoothness and local quality.

\noindent\textbf{Reward Functions.}
We use simple reward functions to penalize invalid masked compositions that could lead to low motion quality, such as excessive smoothness or incoherence. As shown in Table \ref{tab:ablation}, including this reward component is generally beneficial. The reward function is not highly sensitive to reasonable weights $\lambda_k$ around $1.0$. It is particularly important when text alignment needs to be enforced. For example, in Task-3 with the action \emph{raising hand}, the semantic success rate drops significantly without the semantic term. 
We also find that the learned reward model MotionCritic \cite{wangaligning} that aligns with human perception, is not effective for generating highly constrained motions. It tends to assign low scores to rare motions (e.g. side walking).

\begin{figure}[!t]
  \centering
   \includegraphics[width=\linewidth]{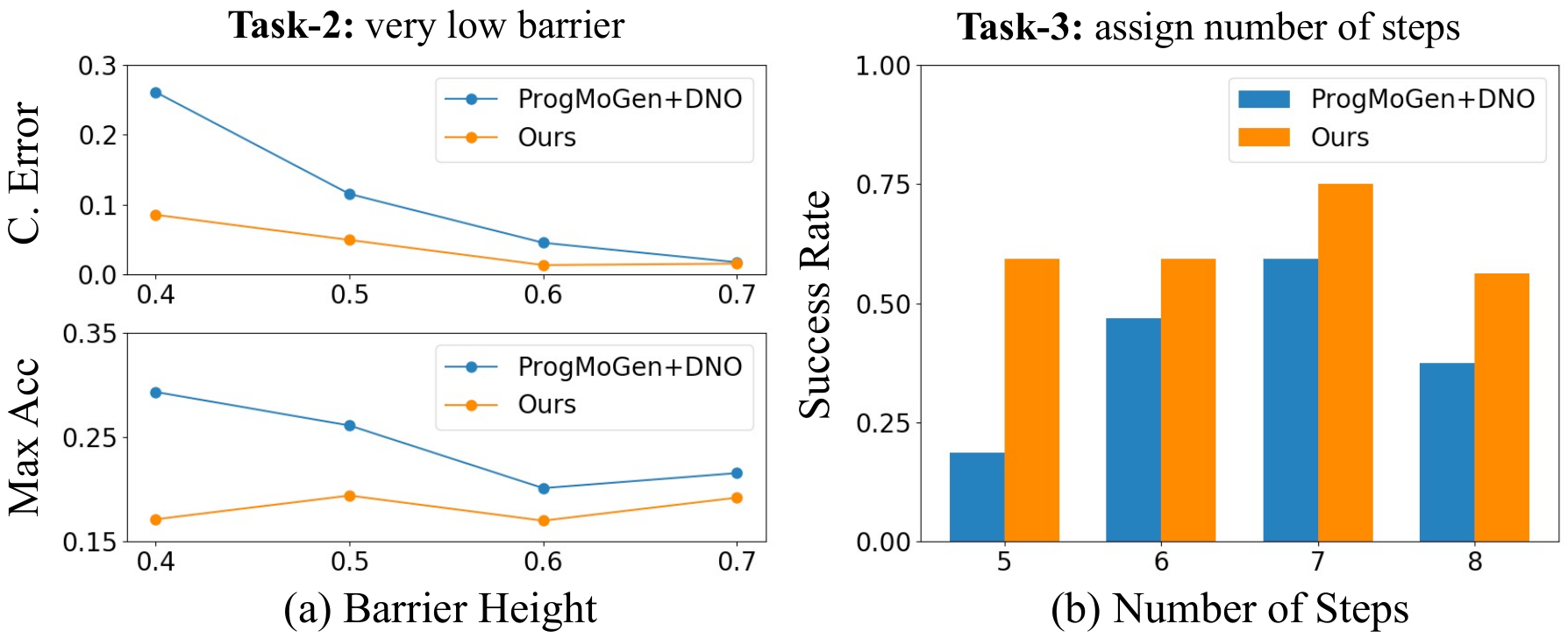}
   \setlength{\abovecaptionskip}{-10pt}
   \caption{Performance on different levels of task difficulty. (a) Different heights of the overhead barrier. (b) Different numbers of steps for the same walking distance. }
   \label{fig:task_difficulty}
\end{figure}

\noindent\textbf{Task Difficulty.} To assess when our method is needed, we increase the difficulty for Task-2 and Task-3 to see where DNO fails. As in Fig.~\ref{fig:task_difficulty}, DNO struggles below 0.5 meters, causing large errors and jitter. Our method handles this height well. In Task-3, DNO struggles with too few or too many steps (e.g., five or eight steps for four meters). Our method improves success rates in all cases.

\subsection{Discussions}
\noindent\textbf{Incorporation with Text Guidance.} 
In Table~\ref{tab:prompt_tuning}, we show that both retrieval-based and text-based guidance can be complementary for solving highly-constrained tasks in certain cases, such as Task-2.
We ask a large language model to adjust the text prompt as an alternative approach for ProgMoGen+DNO. While this improves motion quality, the constraint error remains relatively large, showing that manipulating the prompt alone is insufficient. By combining text guidance with our method, we achieve better overall performance. 
We also observe that the current models still have difficulty generating accurate action counts by simply injecting numbers into prompts.

\begin{figure}[!t]
  \centering
   \includegraphics[width=\linewidth]{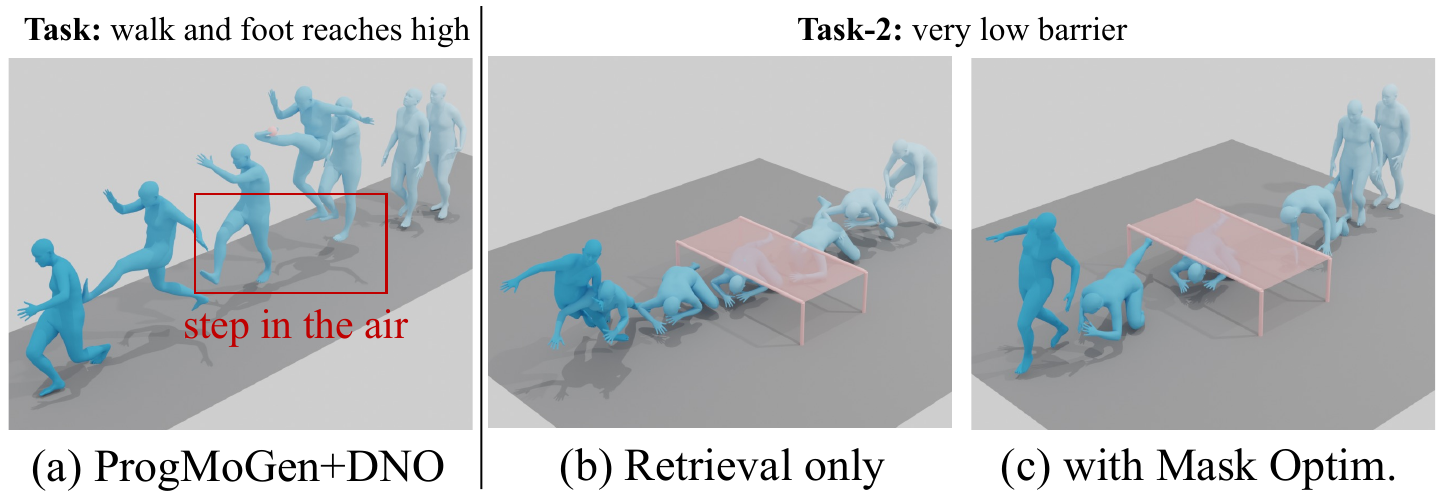}
   \vspace{-0.25in}
   \caption{Qualitative comparison. (a) ProgMoGen+DNO produces unsatisfactory motion for difficult constraints. For Task-2: (b) Using retrieved noise only produces implausible motion when fitting the entire task. (c) We generate plausible motion by combining retrieved noise and random noise with mask optimization.}
   \label{fig:compare_simple_retrieve}
\end{figure}

\noindent\textbf{Diversity.}
 We observe less motion diversity compared to ProgMoGen, as the solution becomes more constrained under very difficult conditions. However, by combining retrieved noise with random noise, we introduce greater diversity beyond a single retrieval. Different motions generated for Task-2 are shown in Fig.~\ref{fig:qualitative} and Fig.~\ref{fig:compare_simple_retrieve} (c).

 \noindent\textbf{Limitations.}
Our method has several phases of optimization and requires around 300 additional optimization steps than DNO. The optimization efficiency could be further improved by training a one-step motion generation model \cite{eyring2024reno}.
Moreover, failure cases of unnatural motion and text misalignment are discussed in the supplementary material.

\setlength{\tabcolsep}{5pt}
\begin{table}[!t]
    \centering
    \small
    \begin{tabular}{lccc} 
    \toprule
   Method            & Foot Skate  & Max Acc.  & C.Err. \\  
   \midrule
   DNO (w. text guidance)   & \textbf{0.154} & 0.194 & 0.000075  \\ 
   Ours            & 0.228 & 0.194 & 0.000049  \\ 
   Ours (w. text guidance)   & 0.161 & \textbf{0.168} & \textbf{0.000027}  \\ 
  \bottomrule
    \end{tabular}
    \caption{Performance on incorporating with text guidance via large language models for Task-2 \textit{very low barrier}. Our method can further improve performance with the help of text guidance.}
    \label{tab:prompt_tuning}
    \vspace{-10pt}
\end{table}
\setlength{\tabcolsep}{1.4pt}

\section{Conclusion}
\label{sec:conclu}

In this paper, we introduce a retrieval-guided diffusion noise optimization framework for highly-constrained human motion generation. The key idea is to retrieve skills from datasets for difficult constraints. We propose relational task parsing, constraint-based retrieval, and masked noise optimization to combine random and retrieved noise, enabling diffusion under complex spatiotemporal and numerical constraints. In the future, we aim to explore LLM reasoning for better task parsing and reward design, and incorporate text guidance to further enhance motion quality.

{\noindent\textbf{Acknowledgements} This work was supported by National Natural Science Foundation of China (Grant No.: 62220106003 and 62572267), Tsinghua University Initiative Scientific Research Program and Tsinghua-Tencent Joint Laboratory for Internet Innovation Technology. }

{
    \small
    \bibliographystyle{ieeenat_fullname}
    \bibliography{main}
}

\clearpage
\setcounter{page}{1}
\maketitlesupplementary

\renewcommand{\thesection}{\Alph{section}}

This supplementary material documents additional implementation details (Section \hyperref[sec:a]{A}), experimental details (Section \hyperref[sec:b]{B}), results and analyses (Section \hyperref[sec:c]{C}), task configuration for quantitative and qualitative evaluation (Section \hyperref[sec:d]{D}) and further discussions  (Section \hyperref[sec:e]{E}) for the proposed method. The video results are also provided in the supplementary material.

\section*{A. Implementation Details}
\label{sec:a}

\begin{figure*}[!t]
  \centering
   \includegraphics[width=0.9\linewidth]{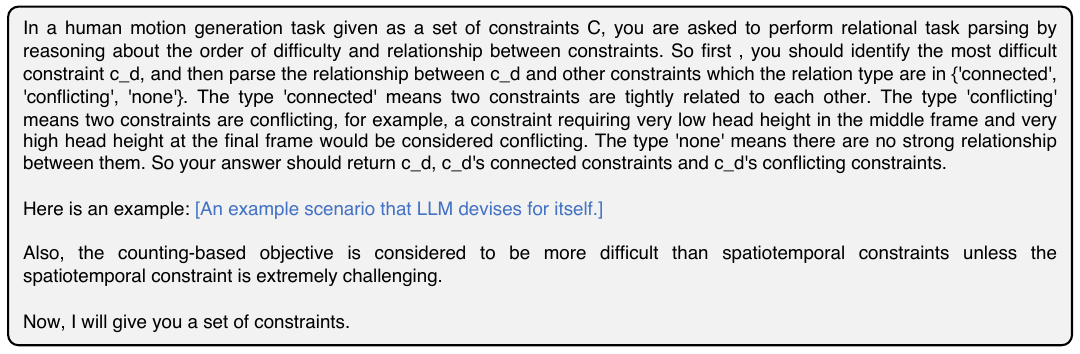}
   \includegraphics[width=0.9\linewidth]{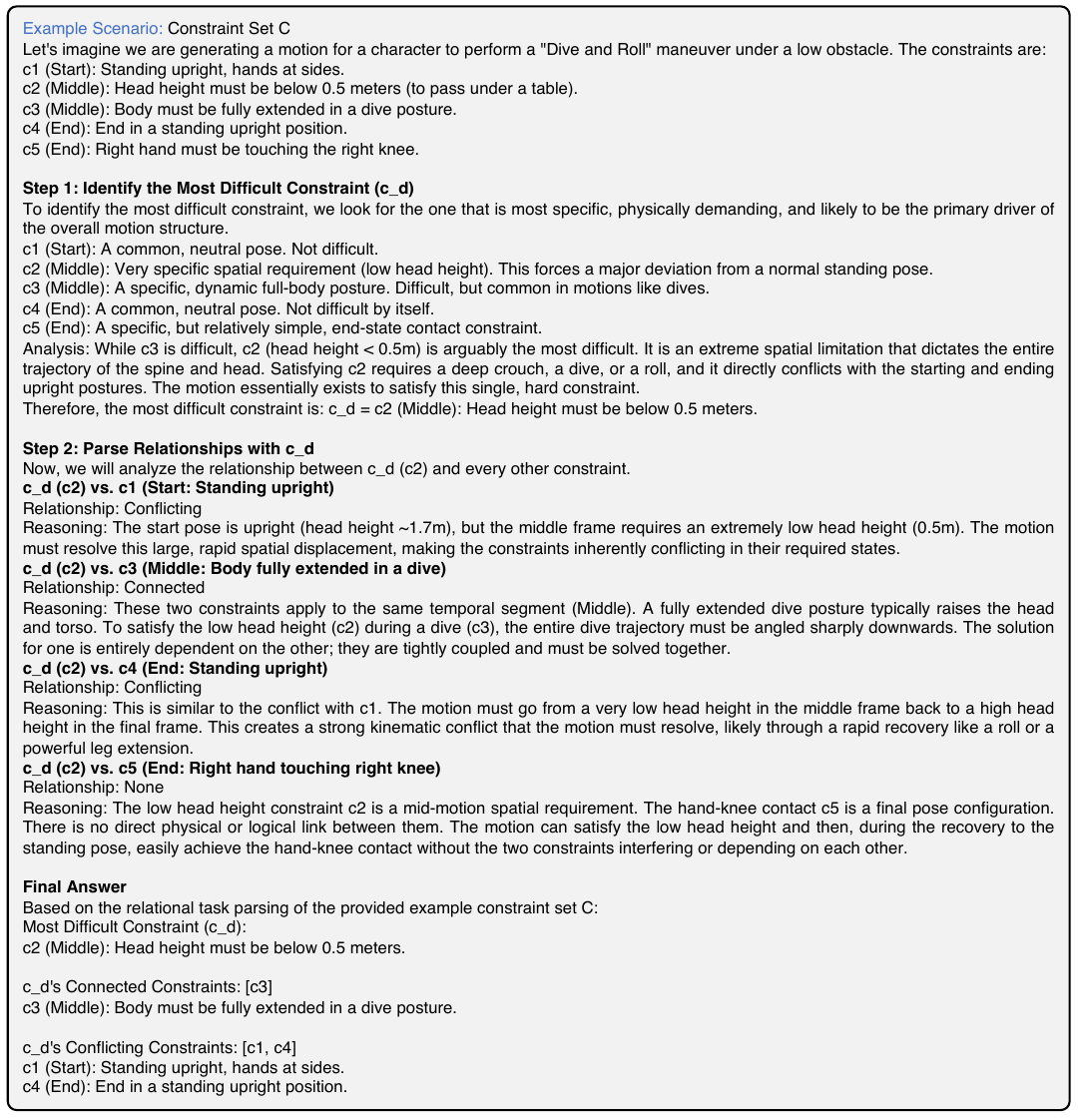}
   \caption{Automatic LLM-based relational task parsing. The above shows the instruction comprising a task description, reasoning rules, and an example case that the LLM devises and reasons itself which is shown in the below.}
   \label{fig:llm_parsing}
\end{figure*}

\noindent\textbf{LLM-based Relational Task Parsing.}
We implement automatic relational task parsing using the large language model DeepSeek-R1 \cite{guo2025deepseek}. The instruction comprises a task format and reasoning rules, along with an example scenario case that the LLM reasons for itself based on the given task description. We feed the textual description or optionally the code implementation of the combined constraint function to the LLM, asking it to identify the difficult constraint $c_D$ and its relationship with other constraints. The instruction is shown in Fig.~\ref{fig:llm_parsing}. The retrieval confidence score is manually specified for the random noise fitting set $C_1$.

\noindent\textbf{Semantic Check.}
We implement semantic check based on keyword compliance. We consider two types of keywords: action verbs and descriptive adverbs for spatial relations. Specifically, we filter out motion samples that (1) contain contradictory adverbs, and (2) lack the specified action verb when the task is defined by numerical constraints. For example, given the prompt ``a man walks forwards'', motion samples with annotation \textit{backwards} are filtered out.

\noindent\textbf{Choices of Masks.}
To determine the mask type, we run Eq.~(\ref{eq:mask_candidate}) for both $2^{N_T}$ temporal masks and $2^{N_S}$ spatial masks, and pick the one with the smallest loss. Given the task description, we can also directly parse the mask type that is more reasonable. Currently, we avoid spatiotemporal masks as they can easily degrade the content preserved in the diffusion noise $z_1$ and $z_2$.

\noindent\textbf{Soft Mask Blending.} To refine sharp boundaries in the binary mask candidate $M$, we further optimize a soft value mask in $sigmoid$ representation. First, we transform the optimized $M$ into an approximate sigmoid form as $M = \sigma (M')$ so that $z'=\sigma (M')z_1 + (1-\sigma (M'))z_2$. We then perform another round of optimization on $M'$ based on Eq.~(\ref{eq:mask_candidate}) in the main paper. In this way, the mask $M$ becomes smoother and its values still lie in $[0,1]$.

\noindent\textbf{Reward Function.}
Eq.~(\ref{eq:reward_func}) in the main paper represents a general form of the reward function and the default value for $\lambda_k$ is set to $1.0$.
Depending on each individual generation task, we can adjust the weight $\lambda_k$ and omit certain terms for simplicity.
In our experiments,
for Task-1, the reward function is designed as
$\mathcal{R} = \mathcal{L}_{\text{jitter}}+\mathcal{L}_{\text{decorr}}+10\mathcal{L}_{\text{foot skate}}$.
For Task-2, the reward function is designed as
$\mathcal{R} = \mathcal{L}_{\text{jitter}}+\mathcal{L}_{\text{decorr}}$.
For Task-3, the reward function is designed as 
$\mathcal{R} = \mathcal{L}_{\text{foot skate}}+0.5\mathcal{L}_{\text{semantic}}$.
Following DNO \cite{karunratanakul2024optimizing}, we adopt a simplified semantic score for \textit{raising hand} as an indicator function for whether the hand height is above a threshold. More precise text-action similarity score can also be adopted.
For the reward function, we use a rectified form of foot skating by defining the ground plane as the lowest foot height in the generated motion sequence.

\begin{figure}[!t]
  \centering
   \includegraphics[width=\linewidth]{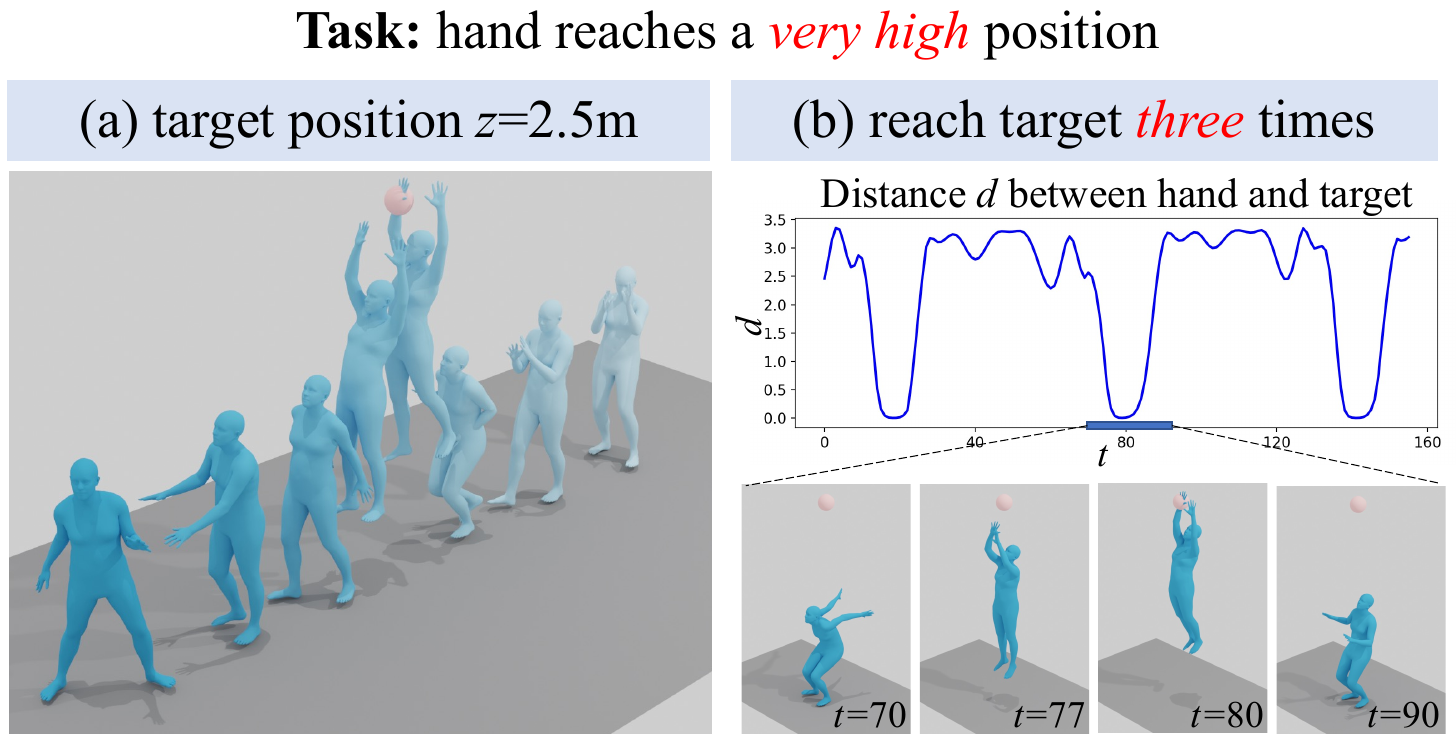}
   \vspace{-10pt}
   \caption{Qualitative examples for Task: \textit{hand reaches a very high position} with different types of constraints. (a) Spatial constraint: the target is located at $z=2.5$ along the walking path. (b) Numerical constraint: the target is located at the origin and the goal is to reach the target position three times.}
   \label{fig:more_results}
\end{figure}

\begin{figure}[!t]
  \centering
   \includegraphics[width=\linewidth]{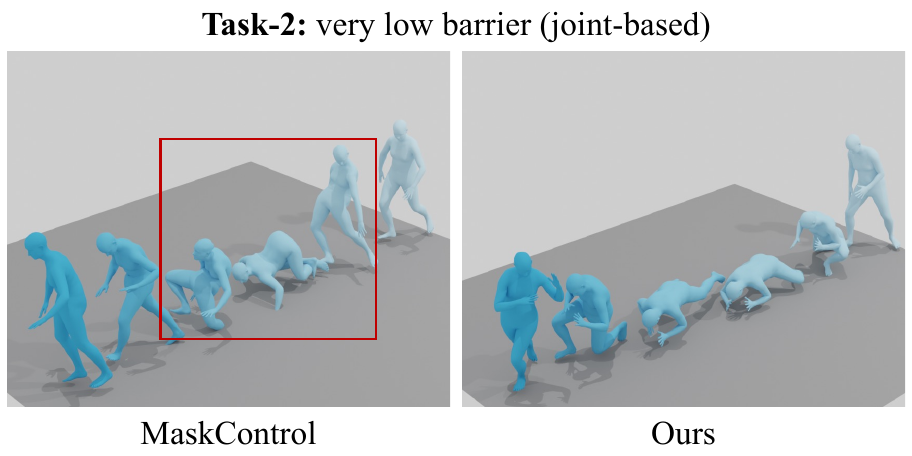}
   \setlength{\abovecaptionskip}{-8pt}
   \caption{Comparison with MaskControl for Task-2 in the joint-based form. MaskControl generates implausible poses given very challenging constraints. The barrier is not visualized for clarity.}
   \label{fig:comp_mask_control}
\end{figure}

\section*{B. Experimental Details}
\label{sec:b}

\textbf{Evaluation Metrics.} 
We calculate the metric of maximum scene penetration for Task-1 and Task-2. 
For Task-1 \textit{very narrow gap}, it is defined as the sum of maximum horizontal penetration depths of the body surface vertices into each wall. For Task-2 \textit{very low barrier}, it is defined as the sum of maximum vertical penetration depths of the body surface vertices into the overhead barrier and the ground.
For Task-3, the success rate is defined as the percentage of samples with the correct number of walking steps. The number of walking steps for the generated motion is calculated by counting rising or falling edges of the foot velocity when crossing a predefined threshold. For semantic success rate, similar to DNO \cite{karunratanakul2024optimizing}, we further check whether the generated motion contains the action \textit{raising hand}. We measure whether there is at least a frame in the generated motion in which the hand is above the shoulder.
For the ablation study in Table 4, the local foot skating is  calculated where the walking distance $z$ is within the range $4<z<5$. The ground height is rectified and the foot height threshold is set to $0.025$ m.

\noindent\textbf{Baseline Details.}
For Unconstrained MDM-RoHM, 
we exclude shape parameters from RoHM representation and derive a 284-dimensional pose feature. The root trajectory is recovered from the relative trajectory representation and the joints and vertices are both recovered from SMPL parameters. We train the motion diffusion model with 600,000 steps with an initial learning rate of $0.0001$. The DDIM-50 model is used for generation.

\section*{C. Additional Experiments and Analyses}
\label{sec:c}

\noindent\textbf{More Qualitative Examples.}
In Fig.~\ref{fig:more_results}, we present results for the task \textit{hand reaches a very high position} under different types of constraints, i.e., spatial and numerical constraints. The numerical constraint for reaching the target three times is implemented as a differentiable function that counts prominent minima in the distance between hand and target, as illustrated in Fig.~\ref{fig:more_results} (b).

\noindent\textbf{More Quantitative Results.}
Previous works \cite{karunratanakul2024optimizing, liu2024programmable} typically report results on very limited handcrafted tasks for evaluating zero-shot constraint-based motion generation. While it is a common challenge to design a comprehensive benchmark for this field, we construct a larger benchmark comprising nine cases derived from this work, covering varied constraint types, task scenes and control joints. The benchmarking tasks include cases in Table~\ref{tab:1}, Table~\ref{tab:2}, Fig.~\ref{fig:qualitative} and Fig.~\ref{fig:more_results}, and further include the task \emph{very high hurdle} in which a human jumps over a high hurdle. The full result is shown in Table~\ref{tab:more_benchmark}. With the proposed retrieval guidance, we achieve lower constraint error on most of the tasks compared to ProgMoGen+DNO.

\noindent\textbf{Effect of Relational Task Parsing.} It is useful for relational task parsing to include difficult constraint $c_D$ with its tightly related constraints into the retrieval set $C_R$. For example, retrieving jointly for the difficult constraint \textit{walk five steps} together with the constraint \textit{walk four meters} yields more effective guidance for optimizing the whole task than retrieving based on \textit{walk five steps} only.
From Fig.~\ref{fig:qualitative} of the main paper, we see that the LLM can well infer the retrieval subset $C_R$ and the target position constraint is correctly identified to be included. In practice, during retrieval, we set a smaller weight for the target position constraint in order not to overshadow the most difficult one $c_D$.

\noindent\textbf{Reward Function Design.} 
We also experiment on MotionCritic \cite{wangaligning} which is a motion quality reward learned from preference alignment. We use $\mathcal{R}=\sigma(-s)$ to transform the reward into loss where $s$ is the reward score and $\sigma$ is the sigmoid function, and set the weight $\lambda$ to $1.0$. The result is shown in Table \ref{tab:appendix_reward_design}. We find that it is not as effective as common heuristic quality check functions for pruning invalid masked noise compositions. 
Also, from Table \ref{tab:appendix_reward_design}, we see that  the design of reward function is not highly sensitive.

\setlength{\tabcolsep}{1.5pt}
\begin{table}[!t]
    \centering
    \scriptsize
    \begin{tabular}{l|lc|cc} 
    \toprule
   Constr. Type & Task Description & Ctrl Joint  & C.Err(DNO) & C.Err(Ours) $\downarrow$  \\  
   \midrule
   \textbf{strict}   & very narrow gap    & full body  & 0.0162 & \textbf{0.0050}   \\
   \textbf{spatial}    & very low barrier & full body          & 0.000115 & \textbf{0.000049} \\ 
       & very high hurdle & full body       & \textbf{0.000} & \textbf{0.000} \\ 
      & reach high position    & hand      & \textbf{0.0013} & 0.0015 \\ 
      & reach high position    & foot      & 0.0013 & \textbf{0.0007} \\ \midrule
   \textbf{numerical}   & walking steps w/ action   & foot      & 0.282 & \textbf{0.0003}  \\
               & number of claps   & hand      &  3.827 & \textbf{0.480} \\
               & number of target touches  & hand  &  9.526 & \textbf{4.190} \\ \midrule
   \textbf{combined} & overhead barrier+walk steps & full body  & 2.250 & \textbf{0.313} \\ 
  \bottomrule
    \end{tabular}
    \caption{Quantitative comparison on the expanded benchmark.}
    \label{tab:more_benchmark}
\end{table}
\setlength{\tabcolsep}{1.4pt}

\setlength{\tabcolsep}{4pt}
\begin{table}[!t]
    \centering
    \small
    \begin{tabular}{lccc} 
    \toprule
    \multicolumn{4}{c}{Task-2: very low barrier} \\
    \toprule
   Reward $\mathcal{R}$            & Foot Skate  & Max Acc.  & C.Err. \\  
   \midrule
   MotionCritic   & 0.254 & 0.217 & 0.000047  \\ 
   $ \mathcal{L}_{\text{jitter}}+\mathcal{L}_{\text{decorr}}$   & 0.228  & 0.194 & 0.000049  \\ 
   $ \mathcal{L}_{\text{jitter}}+\mathcal{L}_{\text{foot skate}}+\mathcal{L}_{\text{decorr}}$   & 0.228 & 0.192 & 0.000059  \\ 
  \bottomrule
    \end{tabular}
    \caption{Comparison on the different reward function design.}
    \label{tab:appendix_reward_design}
\end{table}
\setlength{\tabcolsep}{1.4pt}

\noindent\textbf{Comparison with MaskControl.}
We also conduct qualitative comparison with MaskControl \cite{pinyoanuntapong2024controlmm}. Since MaskControl only supports joint-based constraints, we compare on the joint-based version of Task-2. In this modified task, all joints are constrained to be below $h=0.5$ m to avoid the overhead barrier.
As in Fig.~\ref{fig:comp_mask_control},
MaskControl generates implausible poses when guided by very difficult constraint functions such as a planar barrier. In contrast, our method generates physically plausible and natural poses.

\noindent\textbf{Text Guidance for Numerical Constraints.}
For Task-3, we observe that injecting numbers into prompts does not improve success rates, showing that the current generation models still have difficulty generating accurate action counts.
Recently, AToM \cite{han2025atom} enables control over action frequency with text prompt by finetuning on preference annotations. However, it only supports a limited range of numerical values and cannot handle customized constraints as in our highly-constrained generation tasks.

\noindent\textbf{Analysis on Motion Diversity.}
During masked noise optimization, the mask is optimized to adapt to different randomly sampled noise $z_0$, introducing diversity for the generated motions. 
In the supplementary video results, we show that for the task \textit{very narrow gap}, the generated motions exhibit various local details under the same retrieval, such as speed difference while passing through the gap.

\noindent\textbf{Analysis on Retrieved Motions.}
By minimizing constraint error for the constraint subset $C_R$, the retrieved samples possess certain motion skills for tackling these difficult constraints.
However, the retrieved samples may not necessarily satisfy all the constraints in $C_R$, which still needs to be solved in the final phase of diffusion noise optimization. For example, for the task 
\textit{walk and clap hands wide for four times}, the retrieved motion contains the pattern of clapping four times, but lacks the required motion amplitude (see the visualization in the supplementary video results).

\section*{D. Detailed Task Configuration}
\label{sec:d}

We provide detailed constraint functions for each task in the quantitative evaluation and qualitative examples.
As counting-based numerical constraints (e.g. the number of walking steps) are challenging to design by an end user, following ProgMoGen \cite{liu2024programmable}, we first ask an LLM to provide a reasonable form for the constraint function, and then manually refine it and set appropriate parameters.

\subsection*{D1. Tasks for Quantitative Evaluation}

For \textbf{Task-1} \textit{very narrow gap}, the goal is to pass through a narrow gap formed by two walls and reach the target position. The total constraint function consists of four parts: (1) \texttt{loss\_narrow\_gap} represents a narrow gap of 0.4 meters in width and 3 meters in length, constraining horizontal positions of all body joints and SMPL vertices to the range $-0.2<x<0.2$ when the walking distance is in the range $0.5<z<3.5$; (2) \texttt{loss\_target\_pos} constrains the pelvis joint to reach the target position $z=5$; (3) \texttt{loss\_self\_collision} is a simplified term to avoid collision between arms and the body, constraining the distances between arm/wrist joints and spine joints to be larger than 0.2 m; (4) \texttt{loss\_ground\_contact} constrains the foot to be close to the ground at beginning and end frames. The text prompt is ``a man walks forwards''. The sequence length for generated motions is set to 100 and 150 frames for evaluating on two scenarios of fast and slow walking.

For \textbf{Task-2} \textit{very low barrier}, the goal is to pass through a low overhead barrier and reach the target position. The total constraint function consists of four parts: (1) \texttt{loss\_barrier} represents a barrier of 0.5 meter in height and 1 meter in length, constraining vertical positions of all body joints and SMPL vertices to the range $0<y<0.5$ when the walking distance is in the range $2<z<3$; (2) \texttt{loss\_target\_pos} constrains the pelvis and head joints to reach $z=5$; (3) \texttt{loss\_y\_tstart} constrains the pelvis height to be 0.9 m at the beginning frame and (4) \texttt{loss\_y\_tend} constrains the pelvis height to be 0.9 m at the end frame, encouraging a normal standing pose on both ends. 
The text prompt is ``a man walks forwards''. The motion sequence length is set to 100.

For \textbf{Task-3} \textit{assign number of walking steps}, the goal is to reach the target position with a specified number of steps. The total constraint function consists of three parts: (1) \texttt{loss\_walk\_step} constrains the number of walking steps to be 6; (2) \texttt{loss\_target\_pos} constrains the pelvis joint to reach $z=4$; (3) \texttt{loss\_foot\_both\_ends} constrains both feet to be on the ground with zero velocity at beginning and end frames.  The text prompt is ``a man walks forwards and raises up hands at the same time''. The motion sequence length is set to 100. Specifically,  \texttt{loss\_walk\_step} is implemented as a differentiable peak counting function for foot velocity, where each prominent peak in the smoothed foot velocity signal roughly corresponds to one stepping action. It operates by summing the sigmoid-weighted outputs for all frames identified as prominent peaks:
\begin{align}
\label{eq:step_number}
    \text{step\_number}= \sum_t \sigma &(T(v[t]-v[t+1]))\cdot \\ \nonumber   \sigma&(T(v[t]-v[t-1])) \cdot \\ \nonumber
                 \sigma& (T(v[t]-\theta_v))
\end{align}
where $v$ is the foot velocity, $\sigma$ is the sigmoid function, $\theta_v$ is the velocity threshold and $T$ is the scaling factor. We set a large $T=10000$ to get an accurate step number estimation for comparing against the required number during optimization. We calculate the total number of walking steps on both feet.

\subsection*{D2. Tasks for Qualitative Examples}

Qualitative results for Task-1 and Task-3 are shown in Fig.~\ref{fig:teaser} of the main paper, and a qualitative example for Task-2 is shown in Fig.~\ref{fig:qualitative} of the main paper. 
Details of the remaining tasks from Fig.~\ref{fig:qualitative} of the main paper are provided below.

For the task  \textit{walk with foot reaching a very high position}, the goal is to reach the target destination $z=5$ and also reach a position of 1.8 m high located at $z=2.5$. 
The total constraint function consists of three parts:
(1) \texttt{loss\_reach\_high} constrains one of two feet to reach the high position; 
(2) \texttt{loss\_target\_pos} constrains the pelvis joint to reach $z=5$ at the end frame; 
(3) \texttt{loss\_foot\_contact\_both\_ends} constrains two feet to be on the ground at beginning and end frames. 

For the task \textit{walk and clap hands wide for four times}, 
the goal is to walk to the target position while performing a specified number of hand claps.
The total constraint function consists of four parts: (1) \texttt{loss\_clap\_times} constrains the character to clap its hands four times, which is implemented as a differentiable function that counts the prominent minima of the distance between the hands;
(2) \texttt{loss\_clap\_wide} constrains the inter-hand distance to increase rapidly near the minima point and exceed a threshold;
(3) \texttt{loss\_target\_pos} constrains the pelvis joint to reach the target $z=4$ at the end frame;
(4) \texttt{loss\_hand\_contact} constrains the distance between the hands to be close to zero at prominent minima points.
Specifically, \texttt{loss\_clap\_times} is implemented as a mean square error function to compare groundtruth clap times with the generated motion, and the clap times of the generated motion can be obtained as:
\begin{align}
    \text{clap\_times}= \sum_t \sigma &(T(d[t+1]-d[t]))\cdot \\ \nonumber   \sigma&(T(d[t-1]-d[t])) \cdot \\ \nonumber
                 \sigma& (T(\theta_d-d[t]))
\end{align}
where $d$ is the distance between two hands, $\sigma$ is the sigmoid function, $\theta_d$ is the distance threshold, and $T$ is the scaling factor. We set $T=10000$ for an accurate clapping number estimation.

For the task \textit{walk to avoid overhead barrier in five steps}, the goal is to reach the target position with a specified number of steps and at the same time avoid an overhead barrier of 1.0 m high located at the region of $2<z<3$. The constraint functions are: (1) \texttt{loss\_barrier} constrains all the joints and SMPL vertices to be lower than 1.0 m and above the ground for the walking distance $2<z<3$; (2) \texttt{loss\_walk\_steps} constrains the character to walk for five steps. See its form in Eq. (\ref{eq:step_number}); (3) \texttt{loss\_target\_pos} constrains the pelvis joint to reach the target $z=4$ at the end frame; (4) \texttt{loss\_foot\_both\_ends} constrains the feet to be on the ground at beginning and end frames.

\section*{E. Additional Discussions}
\label{sec:e}

\begin{figure}[!t]
  \centering
   \includegraphics[width=\linewidth]{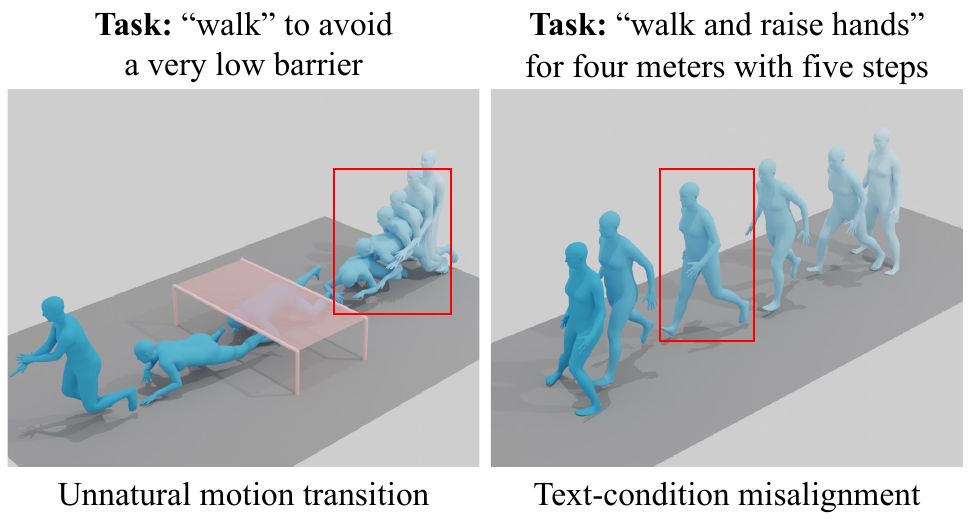}
   \setlength{\abovecaptionskip}{-8pt}
   \caption{Failure cases. Some generated motions of our method may contain unnatural motion transition (left) or may not fully adhere to the text condition (right).}
   \label{fig:failure_cases}
\end{figure}

\noindent\textbf{Limitations and Future Work.}
 For our method, the foot skating remains at a similar level but larger than ProgMoGen+DNO \cite{liu2024programmable, karunratanakul2024optimizing}. 
Also, some generated motions of our method may contain unnatural motion transition as in Fig.~\ref{fig:failure_cases} (left).
 More delicate attentional-layer-based masked noise composition can be designed to further improve motion quality.
Second, the reward guidance is not imposed on the final generated motion, limiting the capability of precise text alignment (see Fig.~\ref{fig:failure_cases} (right)). The semantic alignment term (e.g. VQA score \cite{miao2025noise}) can be included as a constraint function when text alignment should be strictly enforced.
Last, the retrieval configuration involves semantic consistency check, top candidate picking, and coefficients for the retrieval and reward function, which may require some tuning.
As this work introduces a general framework for retrieval-based diffusion noise optimization, its retrieval module can be a focus of future improvement. We can further improve its performance by refining retrieval candidates via re-ranking, using larger in-the-wild motion datasets \cite{fan2025go}, adopting motion clipping and constraint re-writing, and extending to more than one retrieval samples to handle tasks with multiple difficult constraints. While higher-level semantics in motion generation is generally best to be described by text condition itself, one potential application of our method is for abstract actions if they are difficult to be solved by text alone. Such actions can be decomposed into textual and non-textual constraints, where the proposed retrieval guidance can be employed to resolve the challenging non-textual ones.

\noindent\textbf{Differences from other related methods.} 
The motion editing method such as native DNO \cite{karunratanakul2024optimizing} requires a high-quality reference motion, which is not available in the highly-constrained generation task.
STMC \cite{petrovich2024multi} is a purely text-driven spatial and temporal composition method and cannot handle customized constraints.
Different from these approaches, we provide a unified framework for determining what to retrieve and finding the composition of the retrieved noise and random noise.
Moreover, we especially focus on zero-shot goals based on a set of constraint functions, under the training-free framework. For control tasks based on complex joint trajectories or key-frames, methods built upon conditional training, such as MaskControl \cite{pinyoanuntapong2024controlmm}, are more specialized.


\end{document}